\def\year{2022}\relax
\documentclass[letterpaper]{article} 
\usepackage{aaai22}  
\usepackage{times}  
\usepackage{helvet}  
\usepackage{courier}  
\usepackage[hyphens]{url}  
\usepackage{graphicx} 
\usepackage{natbib}  
\usepackage{caption} 
\DeclareCaptionStyle{ruled}{labelfont=normalfont,labelsep=colon,strut=off} 
\usepackage{algorithm}
\usepackage{algorithmic}

\usepackage{bm}
\usepackage{amssymb}
\usepackage{booktabs}
\usepackage{multirow}
\usepackage{amsmath}
\usepackage[switch]{lineno}

\usepackage{newfloat}
\usepackage{listings}
\floatstyle{ruled}
\newfloat{listing}{tb}{lst}{}
\floatname{listing}{Listing}
\title{OneRel: Joint Entity and Relation Extraction with One Module in One Step}
\author{
	Yu-Ming Shang\textsuperscript{\rm 1},
	Heyan Huang\textsuperscript{\rm 1,2}, 
	Xian-Ling Mao\textsuperscript{\rm 1}\thanks{Corresponding author.}\\
}
\affiliations{
	\textsuperscript{\rm 1}School of Computer Science \& Technology, Beijing Institute of Technology, Beijing, China \\
	\textsuperscript{\rm 2}Beijing Engineering Research Center of High Volume Language Information Processing \\ and Cloud Computing Applications, Beijing, China \\
	\{ymshang, hhy63, maoxl\}@bit.edu.cn
}

\begin{document}

\maketitle

\begin{abstract}
	Joint entity and relation extraction is an essential task in natural language processing and knowledge graph construction.
	Existing approaches usually decompose the joint extraction task into several basic modules or processing steps to make it easy to conduct.
	However, such a paradigm ignores the fact that the three elements of a triple are interdependent and indivisible. 
	Therefore, previous joint methods suffer from the problems of cascading errors and redundant information.
	To address these issues, in this paper, we propose a novel joint entity and relation extraction model, named OneRel, which casts joint extraction as a fine-grained triple classification problem.
	Specifically, our model consists of a scoring-based classifier and a relation-specific horns tagging strategy.
	The former evaluates whether a token pair and a relation belong to a factual triple.
	The latter ensures a simple but effective decoding process.
	Extensive experimental results on two widely used datasets demonstrate that the proposed method performs better than the state-of-the-art baselines, and delivers consistent performance gain on complex scenarios of various overlapping patterns and multiple triples.
	
\end{abstract}

\section{Introduction}
	
	\noindent Extracting pairs of entities and their relations in the form of (head, relation, tail) or $(h, r, t)$ from unstructured text is an important task in natural language processing and knowledge graph construction.
	Traditional pipeline approaches \cite{zelenko2003,zhou-2005,chan-roth-2011-exploiting} treat entity recognition and relation classification as two separate sub-tasks. 
	Although flexible, pipeline methods ignore the interactions between the two sub-tasks and are susceptible for the problem of error propagation  \cite{li-ji-2014-incremental}.
	Therefore, recent studies focus on building joint models to obtain entities together with their relations through a unified architecture.
	
	\begin{figure}[t]
		\centering
		\includegraphics[width=1\columnwidth]{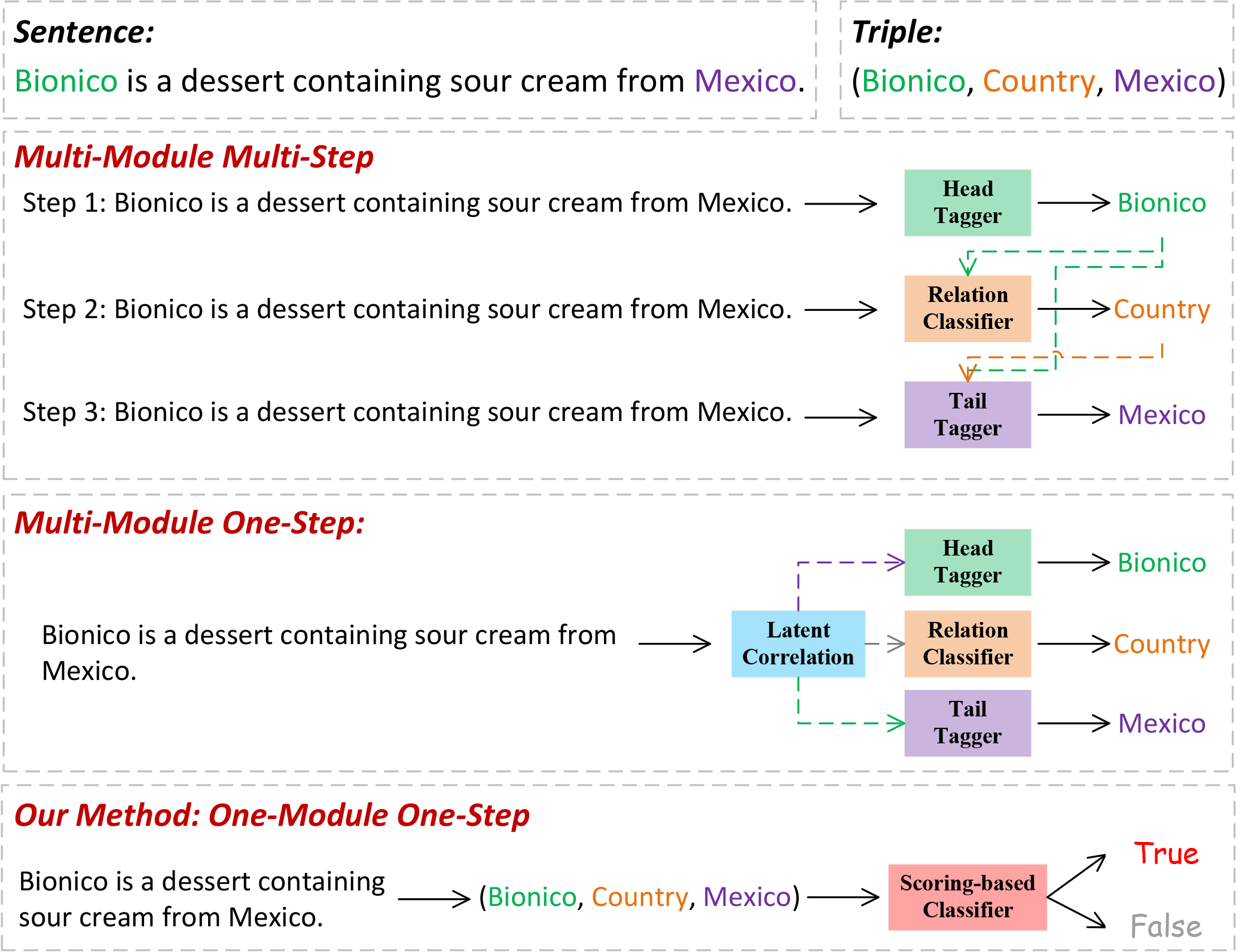} 
		\caption{
		   The extraction processes of existing approaches and our method.
		   The dotted arrow indicates the dependencies between triple elements.
		   Note that there are various extraction paradigms in \textit{multi-module multi-step} approaches, e.g., $(h, t) \rightarrow r$, $h \rightarrow r \rightarrow t$, $r \rightarrow (h, t)$.
		   We just use $h \rightarrow r \rightarrow t$ to illustrate the shortcomings of this kind of methods.
		}
		\label{fig:example}
	\end{figure}

	To make the complex task easy to conduct, existing studies usually decompose the joint extraction into several basic modules or processing steps \cite{bowen2020,zhao-2021-unified}.
	As shown in Figure \ref{fig:example}, 
	according to the extraction procedure of triple elements, these approaches fall into two categories:
	\textit{multi-module multi-step} and \textit{multi-module one-step}.
	The first category utilizes different modules in the framework of cascading classification \cite{fu-etal-2019-graphrel,yuan2020,wei-etal-2020-novel,zheng-2021-prgc,zhao-2021-kbs,zhao-2021-unified} or text generation \cite{zeng-etal-2018-extracting,zeng-2020-CopyMTL,ye-2021-contrasive} to obtain entities and relations step-by-step.
	Although promising, this kind of models suffers from the problem of cascading errors since mistakes in early steps may affect the prediction results of later steps.
	The second category attempts to identify entities and relations separately, and then combine them into triples based on their latent correlations \cite{wang-etal-2020-tplinker,sui2020joint,wang-2021-unified}.
	However, due to insufficient mutual constraints between entities and relations in the separate recognition process, such methods tend to produce redundant information, leading to errors when assembling triples \cite{zheng-etal-2017-joint}.
	
	In fact, the fundamental reason for the above problems is that the decomposition-based paradigm ignores an important property of a triple --- its head entity, relation and tail entity are interdependent and indivisible.
	In other words, it is unreliable to extract one element without fully perceiving the information of the other two elements. 
	To fill this gap, we try to accomplish the joint extraction task from the perspective of triple classification.
	For example, as shown in Figure \ref{fig:example}, ``Bionico" and ``Mexico" are two words in the sentence and \texttt{Country} is a pre-defined relation, all of them are visible in training data. 
	Intuitively, the triple (\textit{Bionico}, \texttt{Country}, \textit{Mexico}) can be directly identified by judging its correctness.
	The idea brings three advantages as follows.
	First, head entity, relation and tail entity are simultaneously fed into one classification module, making it possible to fully capture the dependencies between triple elements, thereby reducing redundant information.
	Second, only one-step classification is used, which is able to effectively avoid the cascading errors.
	Third, the simple architecture of \textit{one-module one-step} empowers the network straightforward and easy to train.
	
	Inspired by the above idea, in this paper, we propose a novel joint entity and relation extraction model, named OneRel, which is capable of extracting all triples from unstructured text with one module in one step. 
	Considering that an entity may consist of multiple tokens, we design a scoring-based classifier and cast the joint extraction task into a fine-grained triple classification problem.
	Specifically, for a token pair $(w_i, w_j)$ and a pre-defined relation $r_k$, the scoring-based classifier measures the correctness of the combination $(w_i, r_k, w_j)$, which will be assigned with a meaningful tag if it is valid and ``-" otherwise.
	To this end, for an input sentence, the output of OneRel is a three-dimensional matrix with each entry corresponding to the classification result of $(w_i, r_k, w_j)$.
	In order to decode entities and relations from the output matrix accurately and efficiently, we introduce a novel relation-specific horns tagging (Rel-Spec Horns Tagging for short) strategy to determine the boundary tokens of head entities and tail entities.
	Experimental results on two widely used benchmark datasets prove that the proposed method outperforms previous approaches and achieves the state-of-the-art performance. 
	
	In summary, our contributions are as follows:
	
	\begin{itemize}
		\item We provide a novel perspective to transform joint extraction into fine-grained triple classification, making it possible to capture the information of head entities, relations and tail entities at the same time.
		
		\item Following our perspective, we introduce a novel scoring-based classifier and a Rel-Spec Horns Tagging strategy. The former is responsible for parallel tagging, and the latter ensures efficient decoding.
		
		\item We evaluate our model on two public datasets, and the results indicate that our method performs better than state-of-the-art baselines, especially for complex scenarios of overlapping triples.
	\end{itemize}

\section{Related Work}
\label{related work}

	Existing joint methods can be roughly divided into two classes according to their extraction procedure of triple elements:

	The first class is \textit{multi-module multi-step}, which uses different modules and interrelated processing steps to extract entities and relations serially. 
	For example,
	a line of works first identify all entities in a sentence, and then perform relation classification between every entity pairs \cite{katiyar-cardie-2017-going,tan-2019,fu-etal-2019-graphrel,liu2020}. 
	The second line of works first detect the relations expressed by a sentence rather than preserve all redundant relations; then head entities and tail entities are predicted \cite{zeng-etal-2018-extracting,yuan2020,zheng-2021-prgc,ma-dual-2021}.
	The third line of works first distinguish all head entities, and then inference corresponding relations and tail entities via sequence labeling or question answering \cite{wei-etal-2020-novel,bowen2020,zhao-2021-kbs,zhao-2021-unified,ye-2021-contrasive}.
	Despite their success, the \textit{multi-module multi-step} methods suffer from the problem of cascading errors, as the mistakes in early steps cannot be corrected in later steps.
	
	The second class is \textit{multi-module one-step}, which extracts entities and relations in parallel, and then combines them into triples.
	For example,
	\citet{miwa-bansal-2016-end,zhang-2017-global,wang-etal-2020-tplinker,wang-2021-unified} treat entity recognition and relation classification as a table-filling problem, where each entry represents the interaction between two individual words.
	\citet{sui2020joint} formulate the joint extraction task as a set prediction problem, avoiding considering the prediction order of multiple triples.
	However, due to insufficient mutual constraints between entities and relations in the separate recognition process, such \textit{multi-module one-step} approaches cannot fully capture the dependencies between predicted entities and relations, resulting in redundant information during triple construction.
	
	Different from existing methods, in this paper, we propose to treat the joint extraction task as a fine-grained triple classification problem, which is able to extract triples from sentences in a \textit{one-module one-step} manner. Therefore, the aforementioned cascading errors and redundant information can be greatly addressed.
	Besides, the classical model Novel-Tagging \cite{zhang-2017-global} designs a complex tagging strategy to establish connections between entities and relations, and can also identify triples from sentences in one-step. Nevertheless, this technique cannot handle overlapping cases because it assumes every entity pair holds at most one relation.

	\begin{figure*}[!ht]
		\centering
		\includegraphics[width=17.5cm]{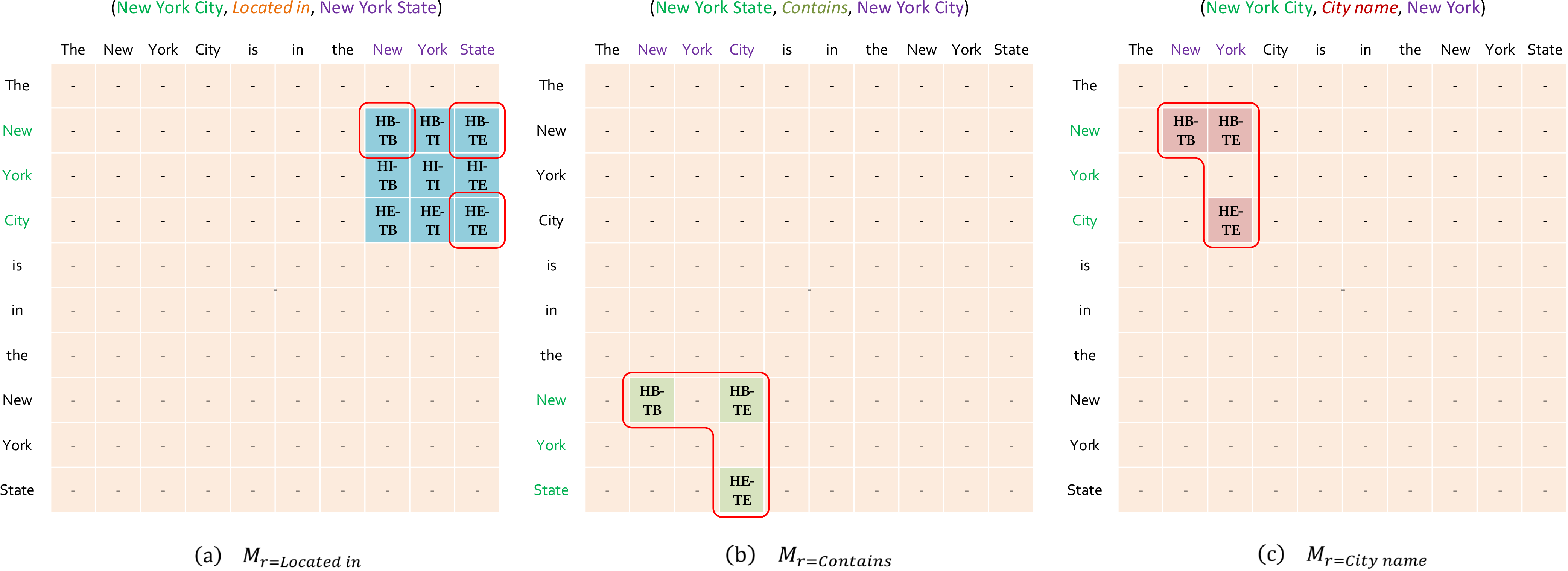} 
		\caption{
			Examples of the Rel-Spec Horns Tagging.
			For the convenience to explanation, we illustrate the sub-matrix with a given relation, e.g., \texttt{Located in}. So, the matrix $\bm{M}$ degenerates into two dimensions, where the rows represent head entities and the columns represent tail entities.
		}
		\label{fig:tagging}
	\end{figure*}

\section{Method}

	In this section, we first give the task definition and notations. 
	Then, we introduce our Rel-Spec Horns Tagging strategy and its decoding algorithm. 
	Finally, we provide a detailed formalization of the scoring-based classifier.

	\subsection{Task Definition}
	
	Given a sentence $\mathcal{S} = \{ w_1, w_2, ..., w_L \}$ with $L$ tokens and $K$ predefined relations $\mathcal{R} = \{ r_1, r_2, ..., r_{K} \}$. The purpose of joint entity and relation extraction is to identify all possible triples $ \mathcal{T} = \{ (h_i, r_i, t_i) \}_{i=1}^{N}$ in $\mathcal{S}$, where $N$ is the number of triples, $h_i, t_i$ are the head entity and tail entity composed of several consecutive tokens, i.e., \texttt{entity.span} $= w_{p:q}$, where $w_{p:q}$ refers to the concatenation of $w_p$ to $w_q$. 

	Note that different triples may share overlapping entities, which poses a big challenge for joint extraction task \cite{zeng-etal-2018-extracting}.

	\subsection{Relation Specific Horns Tagging}

	For a sentence, we design a classifier to assign tags for all possible $(w_i, r_k, w_j)$ combinations, where $w_i, w_j \in \mathcal{S}, r_k \in \mathcal{R}$. 
	We maintain a three-dimensional matrix $\bm{M}^{L \times K \times L}$ to store the classification results (Tagging).
	Therefore, in the test phase, our task is to decode entities and relations from the matrix $\bm{M}$ (Decoding).
	
	\subsubsection{Tagging}
	
	
	We employ the ``BIE" (Begin, Inside, End) signs to indicate the position information of a token in entities. For example, ``HB" means the beginning token of the head entity, ``TE" means the end token of the tail entity.
	As shown in Figure \ref{fig:tagging} (a), for a sentence that expresses the triple (\textit{New York City}, \texttt{Located in}, \textit{New York State}), there are nine special tags (blue tags) in the relation-specific sub-matrix $\bm{M}_{r=Located\; in}$.
	
	According to the insight that an entity can be determined by detecting its boundary tokens \cite{wei-etal-2020-novel}, four types of tags are used in our tagging strategy:
	\textbf{(1) HB-TB.} This tag refers to that both positions are respectively the beginning tokens of a paired head and tail conditioned on a specific relation.
	For example, there is a relation \texttt{Located in} between the two entities ``New York City" and ``New York State", therefore, the classification tag of the combination (``New", \texttt{Located in}, ``New") is assigned with the tag ``HB-TB".
	\textbf{(2) HB-TE.} This tag means that the token corresponding to the row is the beginning of a head entity, meanwhile the token corresponding to the column is the end of a tail entity.
	For instance, ``New" is the start token of ``New York City" and ``State" is the end token of ``New York State", so the combination of (``New", \texttt{Located in}, ``State") is assigned with the tag ``HB-TE". 
	\textbf{(3) HE-TE.} This tag shares a similar logic with ``HB-TB", which means two positions are respectively the end tokens of a paired head entity and tail entity conditioned on a specific relation.
	For example, the combination of (``City", \texttt{Located in}, ``State") is assigned with ``HE-TE".
	\textbf{(4) ``-".} All cells other than the above three cases will be marked as ``-". 
	As we can see from Figure \ref{fig:tagging} (b) and (c), because only the three corners of the rectangle need to be labeled, we vividly name this method as Rel-Spec Horns Tagging.

	Obviously, the tagged matrix $\bm{M}$ is sparse, which has the following advantages:
	First, using three instead of nine special tags can effectively narrow down the potential searching space when conducting classification.
	Second, a sparse $\bm{M}$ means that there are sufficient negative samples during the training process.
	Third, the sparsity of $\bm{M}$ ensures the simplicity and efficiency of triple elements decoding.
	
	Furthermore, our Rel-Spec Horns Tagging can naturally address the complex scenarios with overlapping patterns.
	Specifically, for \textit{EntityPairOverlap} (EPO) case, entity pairs will be labeled in different sub-matrices according to their relations.
	For example in Figure \ref{fig:tagging} (a) and (b), (\textit{New York City}, \texttt{Located in}, \textit{New York State}) and (\textit{New York State}, \texttt{Contains}, \textit{New York City}) are two EPO triples, thus, the two entity pairs are marked in $\bm{M}_{r=Located\;in}$ and $\bm{M}_{r=Contains}$, respectively.
	For \textit{SingleEntityOverlap} (SEO) scenario, if two triples contain the same relation, the two entity pairs will be marked in different parts of $\bm{M}_{r=i}$, otherwise they will be labeled in different sub-matrices according to their relations.
	For the most complicated \textit{HeadTailOverlap} (HTO\footnote{HTO is also called SOO when a triple is represented by (subject, relation, object).}) pattern, e.g. the triple (\textit{New York City}, \texttt{City Name}, \textit{New York}) in Figure \ref{fig:tagging} (c), entity pairs (red tags) are located near the diagonal of $\bm{M}_{r=City\;name}$ and can still be easily decoded.

	\subsubsection{Decoding}
	
	The tagged matrix $\bm{M}^{L \times K \times L}$ marks the boundary tokens of paired head entities and tail entities, as well as the relations between them.
	Therefore, decoding triples from $\bm{M}$ becomes easy and straightforward. 
	That is, for each relation, 
	the spans of head entities are spliced from ``HB-TE" to ``HE-TE";
	the spans of tail entities are spliced from ``HB-TB" to ``HB-TE";
	and two paired entities share the same ``HB-TE".

	\subsection{Scoring-based Classifier}
	
	For an input sentence, we employ a pre-trained BERT \cite{devlin-etal-2019-bert} as sentence encoder to capture the $d$-dimensional token embedding $\bm{e}_i$ for each token:
	\begin{equation}
		\{\bm{e}_1, \bm{e}_2, ..., \bm{e}_L\} = BERT(\{\bm{x}_1, \bm{x}_2, ..., \bm{x}_L\}),
	\end{equation}
	where $\bm{x}_i$ is the input representation of each token. It is the summation over the corresponding token embedding and positional embedding.
	
	Then, we enumerate all possible $(\bm{e}_i, \bm{r}_k, \bm{e}_j)$ combinations and design a classifier to assign high-confidence tags, where $\bm{r}_k$ is the randomly initialized relation representation.
	Intuitively, we can employ a simple classification network whose input is $(\bm{e}_i, \bm{r}_k, \bm{e}_j)$ to achieve this goal.
	However, this intuition has two flaws:
	On the one hand,  a simple classifier is not only unable to fully explore the interactions between entities and relations, but also difficult to model the inherent structural information of triples.
	On the other hand, using $(\bm{e}_i, \bm{r}_k, \bm{e}_j)$ as input means the model needs performing at least $L \times K \times L$ calculations to classify all combinations, which is unacceptable in terms of time.
	
	Inspired by knowledge graph embedding techniques, we borrow the idea from \textsc{HolE} \cite{nickel-2016-hole}, whose score function is defined as:
	\begin{equation}
		f_r(h,t) = \bm{r}^T (\bm{h} \star \bm{t}),
	\end{equation}
	where $\bm{h}$, $\bm{t}$ are head and tail representations, respectively. $\star$ means circular correlation, which is used to mine the latent dependencies between two entities. 
	Here, we redefine the $\star$ operator as a non-linear concatenation projection:
	\begin{equation}
		\bm{h} \star \bm{t} = \phi(\bm{W} [\bm{h};\bm{t}]^T + \bm{b}),
	\end{equation}
	where $\bm{W}\in \mathbb{R}^{d_e \times 2d}$, $\bm{b}$ are trainable weight and bias, $d_e$ denotes the dimension of entity pair representations. $[;]$ is the concatenation operation and  $\phi(\cdot)$ is the ReLU activation function.
	The new definition brings the following benefits:
	First, the score function of our classifier can be seamlessly connected with the output of sentence encoder.
	Second, the mapping function from entity features to entity pair representations can be learned adaptively via the matrix $\bm{W}$.
	Third, the concatenation between two entities is not commutative, i.e., $[\bm{h};\bm{t}] \neq [\bm{t};\bm{h}]$, which is indispensable for modeling asymmetric relations.
	
	Next, we use all relation representations $\bm{R}\in \mathbb{R}^{d_e \times 4 K}$ to simultaneously compute the salience of $(w_i, r_k, w_j)_{k=1}^K$ for a token-pair $(w_i, w_j)$ at once, where 4 is the number of classification tags.
	Therefore, the final score function of our method is defined as:
	\begin{equation}
		\bm{v}_{(w_i, r_k, w_j)_{k=1}^{K}} = \bm{R}^T \phi(\texttt{drop}(\bm{W} [\bm{e}_i;\bm{e}_j]^T + \bm{b})),
		\label{eq:score}
	\end{equation}
	where $\bm{v}$ is the score vector, $\texttt{drop}(\cdot)$ denotes the dropout strategy \cite{nitish-2014-drop} that is used to prevent over-fitting.
	As a result, we achieve the parallel scoring with only two layers of fully connected networks (the matrix $\bm{R}$ can also be regarded as a trainable weight), and reduce the processing steps of actual implementation to $L \times 1 \times L$, even better than TPLinker \cite{wang-etal-2020-tplinker}. 
	Moreover, the score function conforms the idea of \textsc{HolE} and is capable of capturing the correlation and mutual exclusion between relations, which will be verified in the experiments.
	
	Finally, we feed the score vector of $(w_i, r_k, w_j)$ into a softmax function to predict corresponding tags:
	\begin{equation}
		P(\mathrm{y}_{(w_i, r_k, w_j)} | \mathcal{S}) = Softmax(\bm{v}_{(w_i, r_k, w_j)})
	\end{equation}
	
	The objective function of OneRel is defined as:
	\begin{equation}
		\begin{aligned}
			\mathcal{L}_{triple} = &-\frac{1}{ L \times K \times L} \times \\  
			&\sum_{i=1}^{L} \sum_{k=1}^{K} \sum_{j=1}^{L} \log P(\mathrm{y}_{(w_i, r_k, w_j)} = g_{(w_i, r_k, w_j)} | \mathcal{S}),
		\end{aligned}
	\end{equation}
	where $g_{(w_i, r_k, w_j)}$ denotes the gold tag obtained from annotations.

\section{Experiments}
\label{experiment}

	In this section, extensive experiments are conducted to validate the effectiveness of the proposed OneRel and analyze its properties.	
	
	\begin{table*}[t]
		\centering
		\setlength{\tabcolsep}{1.5mm}
		\renewcommand\arraystretch{1.1}
		\begin{tabular}{@{}lcccccccccccccc@{}}
			\toprule[2pt]
			\multicolumn{1}{l}{\multirow{2}{*}{Category}} & \multicolumn{4}{c}{Dataset} & \multicolumn{10}{c}{Details of Test Set}                                          \\ 
			\cmidrule(l){2-5} \cmidrule(l){6-15}
			\multicolumn{1}{l}{}                          & Train    & Valid   & Test   & Relations & Normal & SEO   & EPO   & HTO & N=1   & N=2 & N=3 & N=4 & N\textgreater{}5 & Triples \\ 
			\midrule
			NYT$^*$                                       & 56,195   & 4,999   & 5,000   & 24        & 3,266  & 1,297 & 978   & 45  & 3,244 & 1,045 & 312   & 291  &108 & 8,110   \\
			WebNLG$^*$                                    & 5,019    & 500     & 703    & 171       & 245    & 457   & 26    & 84  & 266   & 171   & 131   &90    & 45 & 1,591   \\
			NYT                                           & 56,195   & 5,000    & 5,000   & 24        & 3,222  & 1,273 & 969   & 117  & 3,240 & 1,047 & 314   & 290  &109 & 8,120   \\
			WebNLG                                        & 5,019    & 500     & 703    & 216       & 239    & 448   & 6     & 85  & 256   & 175   & 138   & 93   & 41 & 1,607   \\ 
			\bottomrule[2pt]
		\end{tabular}
		\caption{Statistics of datasets. $N$ is the number of triples in a sentence.}
		\label{tab:dataset}
	\end{table*}
	
	\subsection{Experimental Settings}
	
	\subsubsection{Datasets and Evaluation Metrics}
	
	Following previous works \cite{wei-etal-2020-novel,wang-etal-2020-tplinker,zheng-2021-prgc}, we evaluate our model and all baselines on two widely used datasets: NYT \cite{riedel2010} and WebNLG \cite{gardent-etal-2017-creating}. 
	The former is generated for distant supervised relation extraction.
	The latter is originally created for natural language generation (NLG).
	Both NYT and WebNLG have two versions: one version only annotates the last word of entities, and the other version annotates the whole span of entities. 
	We denote the first version datasets as NYT$^*$ and WebNLG$^*$, and the second version as NYT and WebNLG.
	To further study the capability of the proposed OneRel in handling complex scenarios, we split the test set by overlapping patterns and triple number. 
	Detailed statistics of the two datasets are described in Table 1.
	
	Three standard evaluation metrics are used in our experiments, i.e., micro Precision (Prec.), Recall (Rec.) and F1-score (F1).
	During evaluation, we adopt \textit{Partial Match} for NYT$^*$ and WebNLG$^*$: an extracted triple (head, relation, tail) is considered to be correct only if the relation and the last word of head and tail are all correct; and use \textit{Exact Match} for NYT and WebNLG: a predicted triple is regarded to be correct only if the whole span of two entities and relation are all exactly matched.
	
	\subsubsection{Implementation Details}
	
	In our experiments, all training process is completed on a work station with an AMD 7742 2.25G CPU, 256G memory, a single RTX 3090 GPU, and Ubuntu 20.04.
	For the pre-trained BERT, we reuse the base cased English model released by Huggingface\footnote{https://huggingface.co/bert-base-cased}, which contains 12 Transformer blocks and the hidden size $d$ is 768.
	We tune our model on the valid set and use grid search to adjust important hyper-parameters.
	Specifically, 
	the batch size is set to 8/6 on NYT/WebNLG, and all parameters are optimized by Adam algorithm \cite{kingma2015} with a learning rate of 1e-5. 
	The dimension of the hidden layer $d_e$ is set to $3 \times d$, the dropout probability in equation (\ref{eq:score}) is 0.1, the max sequence length is set to 100.
	
	\subsubsection{Baselines}
	
	We compare our model with ten state-of-the-art baselines:
	\textbf{GraphRel} \cite{fu-etal-2019-graphrel}, \textbf{RSAN} \cite{yuan2020}, \textbf{MHSA} \cite{liu2020}, \textbf{CasRel} \cite{wei-etal-2020-novel}, \textbf{TPLinker} \cite{wang-etal-2020-tplinker}, \textbf{SPN} \cite{sui2020joint}, \textbf{CGT} \cite{ye-2021-contrasive}, \textbf{CasDE} \cite{ma-dual-2021}, \textbf{RIFRE} \cite{zhao-2021-kbs}, \textbf{PRGC} \cite{zheng-2021-prgc}.

	Note that the sentence encoders used in GraphRel, RSAN and MHSA are LSTM networks, while other baselines employ a pre-trained BERT to obtain feature representations.
	For fair comparison, the reported results for all baselines are directly from the original literature.
	We also conduct ablation test: \textbf{OneRel$^-$} is the model that replaces the classifier of OneRel with $f(w_i, r_k, w_j) = \bm{W}[\bm{e_i};\bm{r_k};\bm{e_j}] + \bm{b}$.

	\subsection{Results and Analysis}

	\subsubsection{Main Results}
	
	Table \ref{tab:main} shows the comparison results of our model against ten baselines on NYT and WebNLG in terms of \textit{Partial Match} and \textit{Exact Match}.
	It can be observed that our method, OneRel, outperforms all the ten baselines and achieves the state-of-the-art F1-score on all datasets.
	Especially on WebNLG$^*$ and WebNLG, OneRel obtains the best performance in terms of all three evaluation metrics, and improves the three indicators on WebNLG to above 90\% for the first time.
	We attribute the outstanding performance of OneRel to its two advantages:
	First, OneRel solves the joint extraction task from the perspective of fine-grained triple classification. Thus, the information of entities and relations can be combined at the same time during extraction, and the redundant information can also be reduced.
	Second, the combination of the scoring-based classifier and the Rel-Spec Horns Tagging accomplishes entity and relation extraction in a straightforward way, effectively avoiding the problem of cascading errors.
	
	Compared with the representative \textit{multi-module multi-step} method PRGC, OneRel achieves 1.3 and 2.5 absolute gain in F1-score on WebNLG$^*$ and WebNLG, respectively.
	This demonstrates that extracting entities and relations simultaneously in one step can effectively address the cascading errors problem.
	Besides, for another \textit{multi-module one-step} model TPLinker which employs four independent tags to detect entities and relations, our OneRel outperforms it by 0.9, 2.4, 0.9 and 4.3 absolute gains on the four datasets, respectively.
	Such results confirm that using one module to extract all triple elements in one step is effective for exploring the interactions between entities and relations.
	These all indicates that \textit{One-module One-step} is expected to become a new paradigm for completing joint extraction task. 
	
	We can also observe that among BERT-based models, OneRel$^-$ achieves a competitive performance with CasRel, TPLinker and CasDE; OneRel obtains the best performance on all datasets. 
	In addition to BERT, OneRel$^-$ and OneRel only use one and two layers of fully connected network, respectively. Their architectures are much simpler than most baselines and easier to train.
	Besides, the performance of OneRel is much better than that of OneRel$^-$ on all datasets, which reveals that it is crucial to capture the dependencies between entities and relations when designing a classifier.
	This conclusion points the way for us to design stronger models in the future.

	\begin{table*}[t]
		\centering
		\setlength{\tabcolsep}{2mm}
		\renewcommand\arraystretch{1.1}
		\scalebox{1}{
				\begin{tabular}{@{}lcccccccccccc@{}}
					\toprule[2pt]
					\multicolumn{1}{c}{\multirow{3}{*}{Model}}  & \multicolumn{6}{c}{\textit{Partial Match}}                    & \multicolumn{6}{c}{\textit{Exact Match}}         \\              
					\cmidrule(l){2-7} \cmidrule(l){8-13}
					\multicolumn{1}{c}{}                       & \multicolumn{3}{c}{NYT$^*$} & \multicolumn{3}{c}{WebNLG$^*$} & \multicolumn{3}{c}{NYT} & \multicolumn{3}{c}{WebNLG} \\ 
					\cmidrule(l){2-4} \cmidrule(l){5-7} \cmidrule(l){8-10} \cmidrule(l){11-13} 
					\multicolumn{2}{c}{}                       Prec.   & Rec.  & F1    & Prec.    & Rec.   & F1     & Prec.  & Rec.   & F1    & Prec.   & Rec.    & F1   \\  
					\midrule
					GraphRel \cite{fu-etal-2019-graphrel}         & 63.9    & 60.0  & 61.9  & 44.7     & 41.1   & 42.9   & -      & -      & -     & -       & -       & -      \\
					RSAN \cite{yuan2020}            & -       & -     & -     & -        & -      & -      & 85.7  & 83.6  & 84.6 & 80.5   & 83.8   & 82.1  \\ 
					MHSA  \cite{liu2020}  & 88.1   & 78.5 & 83.0 &  89.5 & 86.0 & 87.7 & -   & -   & -  & -   &  -    &  -  \\
					\midrule[0.1pt]
					CasRel \cite{wei-etal-2020-novel}          & 89.7    & 89.5  & 89.6  & 93.4     & 90.1   & 91.8   & -      & -      & -     & -       & -       & -      \\
					TPLinker \cite{wang-etal-2020-tplinker}        & 91.3    & 92.5  & 91.9  & 91.8     & 92.0   & 91.9   & 91.4   & 92.6   & 92.0  & 88.9    & 84.5    & 86.7   \\
					SPN \cite{sui2020joint}           & 93.3    & 91.7  & 92.5  & 93.1     & 93.6   & 93.4   & 92.5   & 92.2   & 92.3  & -       & -       & -      \\
					CGT  \cite{ye-2021-contrasive}            & \textbf{94.7}    & 84.2  & 89.1  & 92.9     & 75.6   & 83.4   & -      & -      & -     & -       & -       & -      \\
					CasDE \cite{ma-dual-2021}    & 90.2    & 90.9  & 90.5  & 90.3     & 91.5   & 90.9   & 89.9   & 91.4   & 90.6  & 88.0    & 88.9    & 88.4   \\
					RIFRE  \cite{zhao-2021-kbs}         & 93.6    & 90.5  & 92.0  & 93.3     & 92.0   & 92.6   & -      & -      & -     & -       & -       & -      \\
					PRGC  \cite{zheng-2021-prgc}           & 93.3    & 91.9  & 92.6  & 94.0     & 92.1   & 93.0   & \textbf{93.5}   & 91.9   & 92.7  & 89.9    & 87.2    & 88.5   \\ 
					\midrule
					OneRel$^-$           & 91.3    & 90.5  & 90.9  & 93.8   & 91.4 & 92.6   &  91.1  & 90.4 & 90.8 & 90.5   & 88.2 & 89.4  \\ 
					OneRel           & 92.8    & \textbf{92.9}  & \textbf{92.8}  & \textbf{94.1}     & \textbf{94.4}   & \textbf{94.3}   & 93.2  &  \textbf{92.6}  & \textbf{92.9} & \textbf{91.8}    & \textbf{90.3}    & \textbf{91.0}   \\ 
					\bottomrule[2pt]
				\end{tabular}
		}
		\caption{Precision(\%), Recall (\%) and F1-score (\%) of our proposed OneRel and baselines.}
		\label{tab:main}
	\end{table*}

\subsubsection{Detailed Results on Complex Scenarios}

	\begin{table*}[t]
		\setlength\tabcolsep{1mm}
		\renewcommand\arraystretch{1.2}
		\centering
		\scalebox{1}{
			\begin{tabular}{@{}lcccccccccccccccccc@{}}
				\toprule[2pt]
				\multicolumn{1}{c}{\multirow{2}{*}{Model}} & \multicolumn{9}{c}{NYT$^*$}                                                   & \multicolumn{9}{c}{WebNLG$^*$}                                                \\ 
				\cmidrule(l){2-10} \cmidrule(l){11-19} 
				\multicolumn{1}{c}{}                       & Normal & EPO  & SEO  & HTO & N=1  & N=2  & N=3  & N=4  & N$\geq$5 & Normal & EPO  & SEO  & HTO & N=1  & N=2  & N=3  & N=4  & N$\geq$5 \\ 
				\midrule
				CasRel                                     & 87.3   & 92.0 & 91.4 &  77.0$^\S$   & 88.2 & 90.3 & 91.9 & 94.2 & 83.7         & 89.4   & 94.7 & 92.2 &  90.4$^\S$ & 89.3 & 90.8 & 94.2 & 92.4 & 90.9             \\
				TPLinker                                   & 90.1   & 94.0 & 93.4 & 90.1$^\S$  & 90.0 & 92.8 & 93.1 & 96.1 & 90.0             & 87.9   & 95.3 & 92.5 &  86.0$^\S$ & 88.0 & 90.1 & 94.6 & 93.3 & 91.6             \\
				SPN                                      & 90.8   & 94.1 & 94.0 &  -  & 90.9 & 93.4 & \textbf{94.2} & 95.5 & 90.6                & -     &  -   &    -  &   - & 89.5 & 91.3 & \textbf{96.4} & 94.7 & 93.8             \\
				PRGC                                       & \textbf{91.0}   & 94.5 & 94.0 &  81.8  & \textbf{91.1} & 93.0 & 93.5 & 95.5 & 93.0           & 90.4   & \textbf{95.9} & 93.6 &  94.6 & 89.9 & 91.6 & 95.0 & 94.8 & 92.8             \\
				\midrule
				OneRel                                   & 90.6   & \textbf{95.1}  & \textbf{94.8} &   \textbf{90.8}  & 90.5 & \textbf{93.4} & 93.9 & \textbf{96.5} & \textbf{94.2}         &  \textbf{91.9} & 95.4  & \textbf{94.7}  & \textbf{94.9} & \textbf{91.4} & \textbf{93.0}   & 95.9 & \textbf{95.7} & \textbf{94.5}                \\ 
				\bottomrule[2pt]
			\end{tabular}
		}
		\caption{F1-score (\%) on sentences with different overlapping patterns and different triple numbers. $\S$ marks the results reported by \cite{zheng-2021-prgc}.}
		\label{tab:type}
	\end{table*}
	
	To verify the ability of our method in handling overlapping patterns and multiple triples, we conduct two extended experiments on different subsets of NYT$^*$ and WebNLG$^*$.
	We select four powerful models as baselines and the detailed results are shown in Table \ref{tab:type}.

	It can be observed that our OneRel achieves the best F1-score on 13 of the 18 subsets, especially for the most complex cases \textit{HTO} and $N\geq5$.
	The HTO pattern contains two situations: one is nested entities that most previous approaches cannot accurately identify, e.g., the triple  (\textit{Bruce Lee}, \texttt{Family name}, \textit{Lee}). The other is that the head entity and tail entity share the same words, for example, the sentence ``\textit{Native Americans in the United States are one of the ethnic groups of the country.}" in WebNLG$^*$ expresses the triple  (\textit{States}, \texttt{ethnicGroup}, \textit{States}). 
	Besides, a $N\geq5$ sentence may contain SEO, EPO and HTO patterns at the same time, which brings a big challenge to existing approaches.
	Nevertheless, our OneRel achieves the best performance on both HTO and $N\geq5$ of NYT$^*$ and WebNLG$^*$,
	which adequately proves that our Rel-Spec Horns Tagging is able to address the overlapping triples problem from design, and more robust than baselines when dealing with the complicated scenarios.

\subsubsection{Results on Different Sub-tasks}
	
	\begin{table}[t]
		\centering
		\setlength\tabcolsep{1.5mm}
		\renewcommand\arraystretch{1}
		\begin{tabular}{@{}cccccccc@{}}
			\toprule[2pt]
			\multirow{2}{*}{Model}  & \multirow{2}{*}{Element} & \multicolumn{3}{c}{NYT$^*$} & \multicolumn{3}{c}{WebNLG$^*$} \\ 
			\cmidrule(l){3-5} \cmidrule(l){6-8}
			&                          & Prec.   & Rec.  & F1    & Prec.    & Rec.   & F1     \\ \midrule
			\multirow{3}{*}{CasRel} & $(h, t)$& 89.2    & 90.1  & 89.7  & 95.3     & 91.7   & 93.5   \\
			& $r$ & 96.0    & 93.8  & 94.9  & 96.6     & 91.5   & 94.0   \\
			& $(h, r, t)$                  & 89.7    & 89.5  & 89.6  & 93.4     & 90.1   & 91.8   \\ \midrule
			\multirow{3}{*}{SPN}  & $(h, t)$  & 93.2    & 92.7  & 92.9  & 95.0     & 95.4   & 95.2   \\
			& $r$   & 96.3    & 95.7  & 96.0  & 95.2     & 95.7   & 95.4   \\
			& $(h, r, t)$          & 93.3    & 91.7  & 92.5  & 93.1     & 93.6   & 93.4   \\ \midrule
			\multirow{3}{*}{PRGC}  & $(h, t)$  & \textbf{94.0}    & 92.3  & 93.1  & 96.0     & 93.4   & 94.7   \\
			& $r$   & 95.3    & 96.3  & 95.8  & 92.8     & 96.2   & 94.5   \\
			& $(h, r, t)$          & \textbf{93.3}    & 91.9  & 92.6  & 94.0     & 92.1   & 93.0   \\ \midrule
			\multirow{3}{*}{OneRel} & $(h, t)$  & 93.3    & \textbf{93.4}  & \textbf{93.3}  & \textbf{96.2}    & \textbf{96.5}   & \textbf{96.3}\\
			& $r$                        & \textbf{96.7}    & \textbf{96.9}  & \textbf{96.8}  & \textbf{96.7}     & \textbf{97.0}   & \textbf{96.8}   \\
			& $(h, r, t)$      & 92.8    & \textbf{92.9}  & \textbf{92.8}  & \textbf{94.1}     & \textbf{94.4}   & \textbf{94.3}   \\ 
			\bottomrule[2pt]
		\end{tabular}
	\caption{Results on triple elements. $(h, t)$ denotes the entity pair and $r$ means the relation.}
	\label{tab:subtask}
	\end{table}

	We further explore the performance of OneRel on different sub-tasks, i.e., entity pair recognition and relation classification.
	From Table \ref{tab:subtask}, it can be found that our OneRel outperforms all the baselines on most test instances of NYT$^*$ and all the indicators of WebNLG$^*$. These encouraging results once again verifies our motivation. 
	
 	Interestingly, the four models show the same trend on the two datasets:
	for NYT$^*$, there is an obvious gap between the F1-score on $(h, t)$ and $r$;
	for WebNLG$^*$, the performance on $(h, t)$ and $r$ are much higher than that of $(h, r, t)$.
	Based on this phenomenon, previous researchers have analyzed that entity pair recognition and triple formation are two bottlenecks of the joint extraction task  \cite{sui2020joint}.
	We believe that in addition to the above reasons, the characteristic of the datasets is also an important factor.
	Concretely,
	NYT$^*$ contains lots of EPO triples, which means that the impact of wrongly recognized entity pairs is much greater than the influence of mistakenly classified relations.
	Suppose a sentence that expresses three triples (\textit{Obama}, \texttt{President of}, \textit{United States}),  (\textit{Obama}, \texttt{Live in}, \textit{United States}),  (\textit{Obama}, \texttt{Place of birth}, \textit{United States}).
	If one relation is wrongly classified, the recall on $r$ may be 0.67.
	While if one entity pair is incorrectly recognized, the recall on $(h, t)$ may drop to 0. 
	In contrast, the proportion of EPO patterns in WebNLG$^*$ is much smaller than that of NYT$^*$ (3.6\% vs 19.6\%), thus, the performance of models on entity pair recognition is consistent with that on relation classification.
	Besides, WebNLG$^*$ contains more relations than NYT$^*$ (171 vs 24) and some of them are confusing, e.g., \texttt{Leader} and \texttt{LeaderName}.
	So, the triple formation on WebNLG$^*$ is more difficult than NYT$^*$.
	
\subsubsection{Model Efficiency}

	\begin{table}[]
		\centering
		\setlength\tabcolsep{0.85mm}
		\renewcommand\arraystretch{1}
		\begin{tabular}{@{}ccccc@{}}
			\toprule[2pt]
			Dataset                 & Model    & Training Time & Inference Time & F1   \\ \midrule
			\multirow{2}{*}{NYT$^*$}    & TPLinker & 1592   & 46.2           & 91.9 \\
			& OneRel   &    \textbf{1195}         & \textbf{41.5}           & \textbf{92.9} \\ 
			\midrule
			\multirow{2}{*}{WebNLG$^*$} & TPLinker &      599    & \textbf{40.1}           & 91.9 \\
			& OneRel   &        \textbf{88}     & 45           & \textbf{94.3} \\ 
			\bottomrule[2pt]
		\end{tabular}
		\caption{Comparison of the model efficiency. Training Time (s) means the time required to train one epoch, Inference Time (ms) is the time to predict triples of one sentence.}
		\label{tab:efficiency}
	\end{table}
	
	We evaluate the model efficiency with respect to \textit{Training Time} and \textit{Inference Time} of the most similar baseline TPLinker in two datasets NYT$^*$ and WebNLG$^*$, and the results are shown in Table \ref{tab:efficiency}.
	In this experiment, the batch size of the two models during training and testing are set to 6 and 1, respectively, and the max length of input sentence is 100.	
	Although the theoretical complexity of the two models is $\mathcal{O}({KL^2})$, OneRel is better than TPLinker in terms of parallel processing, i.e., OneRel processes $K$ relations at a time, while TPLinker handles one relation at each step.
	Therefore, when the size of relations increases from 24 (NYT$^*$) to 171 (WebNLG$^*$), the training time of OneRel increases from 1.3$\times$ faster than TPLinker to an astonishing 6.8$\times$.
	As opposed to what we observed in \textit{Training Time}, the F1-score of OneRel is much better than that of TPLinker.
	This confirms the efficiency  and the learning ability of our proposed classifier.
	Besides, the inference time of OneRel is similar to that of TPLinker on NYT$^*$ and WebNLG$^*$, which illustrates the effectiveness and rationality of our proposed Rel-Spec Horns Tagging.
	
\subsubsection{Topology Structure of Relations}
	
\begin{figure}[t]
		\centering
		\includegraphics[width=1\columnwidth]{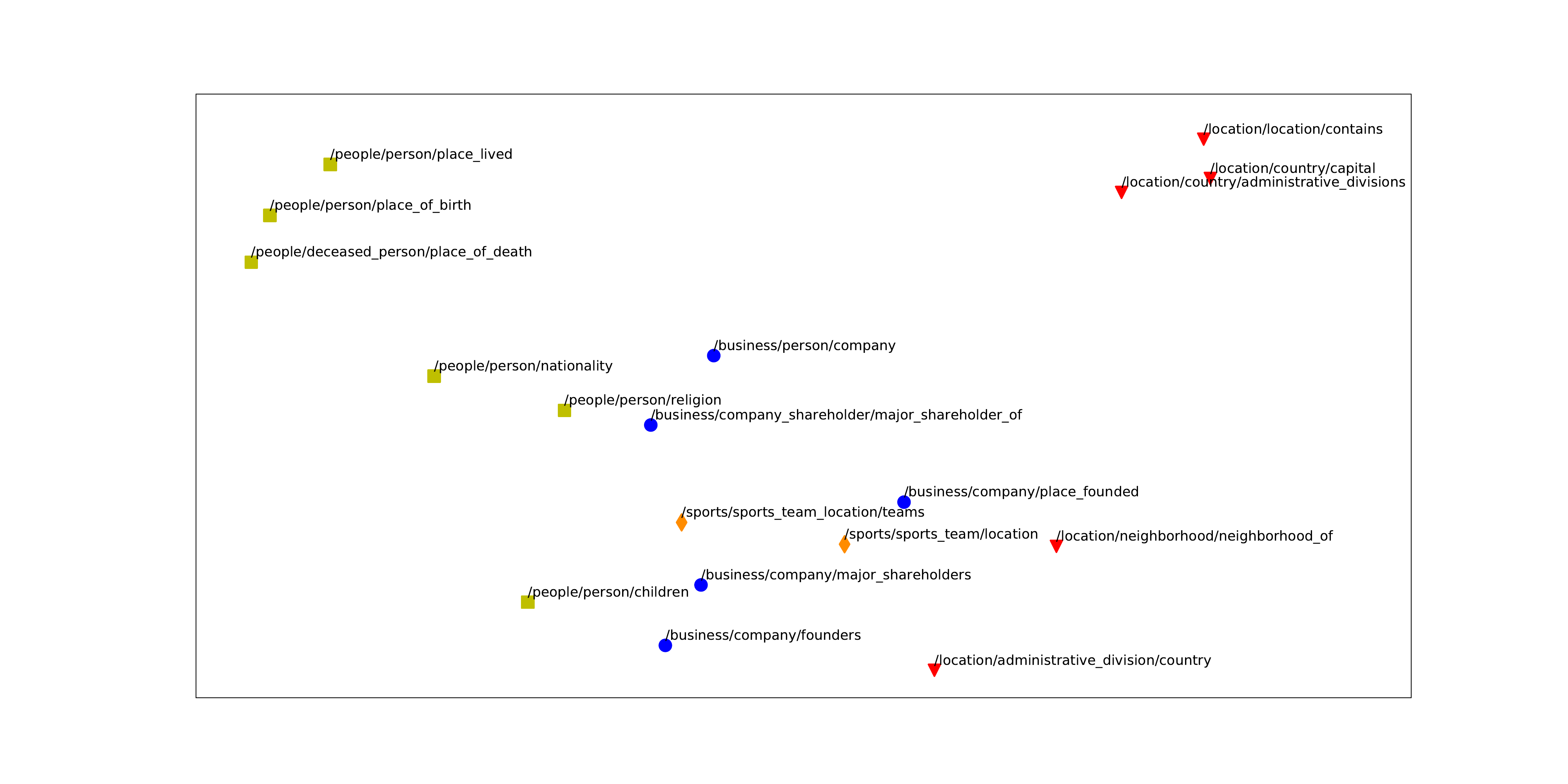} 
		\caption{
			(Best viewed in color and zoom in.) Visualization of relations on NYT dataset.
		}
		\label{fig:visualization}
	\end{figure}

	Our scoring-based classifier borrows the idea of \textsc{HolE} \cite{nickel-2016-hole}, and theoretically, it should also be able to learn the correlation and mutual exclusion between relations.
	To verify the learning ability of our classifier, we visualize the relation representations of NYT by the $t-$SNE \cite{maaten2008visualizing}, which is a nonlinear dimensionality reduction algorithm. We omit 6 long-tail relations that appear less than 50 times in whole training set, and the results are shown in Figure \ref{fig:visualization}.
	It can be observed that
	the topology structure of relations reflects their inherent connections.
	For instance, the relations related to \texttt{people} are on the left region of the figure, while the relations related to \texttt{location} are on the right.
	Especially, the spatial positions of \texttt{place\_lived}, \texttt{place\_of\_birth} and \texttt{place\_of\_death} are very close, which is in consistent with our common sense.
	This property is critical for predicting EPO triples. That is, if the model predicts the relation \texttt{place\_of\_birth} for an entity pair, we can infer that the entity pair may hold another relation \texttt{place\_lived} with a high-probability. 
	These all suggest that our method has learned not only the features of a specific dataset, but also the general knowledge that conforms to the real world.
	Thus, our OneRel may have generalization capabilities.

\section{Conclusion}

	In this paper, we provide a novel perspective to transform the joint extraction task into a fine-grained triple classification problem, and propose a novel joint model with a scoring-based classifier and Rel-Spec Horns Tagging strategy to obtain triples with one module in one step, which greatly alleviates the problems of cascading errors and redundant information. Experiments on public datasets show that our model performs better than the state-of-the-art approaches on different scenarios.	

	In the future, we would like to explore the following directions:
	\begin{itemize}

		\item To improve the efficiency of the model, we design a simplified version of \textsc{HolE} as the score function. We will next try to design a more efficient and powerful score function to further strengthen its ability of capturing the connections between entities and relations.
		
		\item We would like to explore the idea of triple classification in other information extraction problems, such as event extraction.
	\end{itemize}

\section{Acknowledgments}

	The work is supported by National Key R\&D Plan (No. 2018YFB1005100), National Natural Science Foundation of China (No. 61751201, 61772076, and 61732005), Natural Science Fund of Beijing (No. Z181100008918002),  Central Leading Local Project(No. 2020L3024), BDAS National Engineering Laboratory Open Project (No. CASNDST202006) and Fujian Provincial DSTLST Project (No. 2019H0026).

\bibliographystyle{aaai22}
\bibliography{aaai22}

\end{document}